\documentclass[11pt]{article}

\usepackage{eacl}

\usepackage{inconsolata}

\usepackage{times}
\usepackage{latexsym}

\usepackage[T1]{fontenc}

\usepackage[utf8]{inputenc}

\usepackage{microtype}

\usepackage{inconsolata}

\usepackage{amsmath}
\usepackage{amssymb}
\usepackage{amsthm}
\usepackage{graphicx}
\usepackage{mathtools}
\usepackage{listings}
\usepackage{tikz}
\usepackage{siunitx}
\usepackage{tipa}
\usepackage{float}
\usepackage{bbm}
\usepackage{breqn}

\usepackage{caption}
\usepackage{subcaption}

\usepackage{todonotes}

\makeatletter
\newcommand*\iftodonotes{\if@todonotes@disabled\expandafter\@secondoftwo\else\expandafter\@firstoftwo\fi}  
\makeatother

\usepackage{booktabs, multirow} 

\title{A Discerning Several Thousand Judgments:\\GPT-3 Rates the Article + Adjective + Numeral + Noun Construction}

 \author{Kyle Mahowald \\
        \texttt{mahowald@utexas.edu} \\
         Department of Linguistics\\
         The University of Texas at Austin}

\begin{document}

\maketitle

\begin{abstract}
Knowledge of syntax includes knowledge of rare, idiosyncratic constructions. 
LLMs must overcome frequency biases in order to master such constructions. In this study, I prompt GPT-3 to give acceptability judgments on the English-language Article + Adjective + Numeral + Noun construction (e.g., ``a lovely five days''). I validate the prompt using the CoLA corpus of acceptability judgments and then zero in on the AANN construction. I compare GPT-3's judgments to crowdsourced human judgments on a subset of sentences. GPT-3's judgments are broadly similar to human judgments and generally align with proposed constraints in the literature but, in some cases, GPT-3's judgments and human judgments diverge from the literature and from each other.
\end{abstract}

\section{Introduction}

Consider the English Article + Adjective + Numeral + Noun (AANN) construction: `` \textbf{a beautiful 228 pages} [iWeb]'' or ``The president has had \textbf{a terrible five weeks} [COCA]''.
Usually cardinal numbers precede the adjective (``five terrible weeks''), but here the adjective precedes the numeral.
More strangely, the normally singular article ``a'' in this case is followed by a plural noun phrase.\footnote{Data and code: \url{https://github.com/mahowak/aann-public/}}

\begin{figure}[t]
    \centering
    \includegraphics[width=.9\columnwidth]{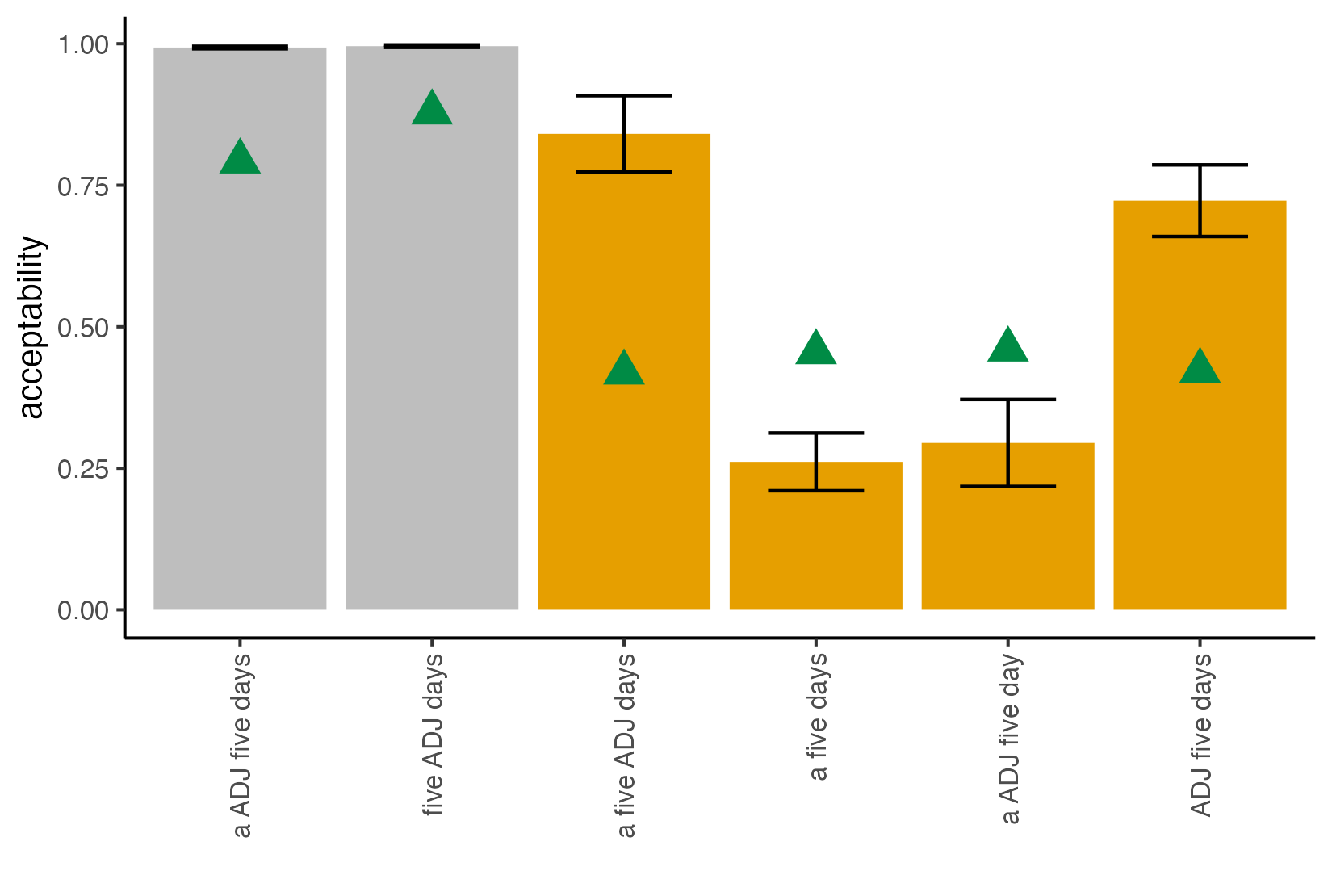}
    \caption{GPT-3 acceptability judgments (bars), compared to human ratings (green triangles) on a matched set of sentences. The comparison is between the AANN construction and the standard alternative, as well as 4 degenerate versions. Both humans and GPT-3 rank the AANN construction as being as acceptable as the standard, and all degenerate constructions are rated significantly lower.}
    \label{fig:fig1}
\end{figure}

An eclectic dozen or so papers have been written on the construction, many focused on elucidating relevant semantic and syntactic constraints  \citep{goldberg2017one,jackendoff_x_1977,dalrymple_amazing_2019,bylinina_zero_2018,ionin_cardinals_2018,ionin_singular_2004,solt_two_2007,keenan_pleasant_2013}.
The presence of the modifier is crucial: ``a 228 pages'' is unacceptable.
The type of modifier is also crucial: ``a pink 228 pages'' is odd because color words are ``stubbornly distributive'' \citep{schwarzschild_stubborn_2011} and thus cannot refer to a set of items as a whole: the nominal phrase needs to function as a unit \citep{solt_two_2007}.
These idiosyncratic constraints are typical of \textit{constructions} \citep{goldberg2019explain}.

Prior work on the AANN construction has focused on characterizing the semantic and syntactic constraints on the construction and proposing analyses in various frameworks. For instance, 
\citet{solt_two_2007} focuses on how the construction coerces the phrase into a singular noun phrase; \citet{keenan_pleasant_2013} proposes treating it as akin to a partitive, and \citet{dalrymple_amazing_2019} give an LFG analysis.

The same properties that make AANN interesting from the perspective of human language use---its low frequency but high sensitivity to constraints---also make it interesting from the perspective of LLMs and what they learn about linguistic structure.
Indeed, much work on LLM syntactic competence has centered on ubiquitous abstract features of grammar like subject-verb number agreement \citep[e.g.,][]{linzen_assessing_2016,gulordava-etal-2018-colorless}, part of speech \citep{tenney-etal-2019-bert}, and syntactic dependencies \citep[e.g.,][]{hewitt_structural_2019}.
That said, there has also been a recent spate of research on construction-grammar-inspired approaches in NLP, including studies showing that LLMs have access to construction information \citep{tayyar-madabushi-etal-2020-cxgbert,tseng-etal-2022-cxlm,weissweiler2022better} and capture verb argument construction biases \citep{hawkins-etal-2020-investigating} as well as fine-grained lexical semantic information \citep{petersen2022lexical}. Moreover, sentences with similar constructions cluster in embedding space \citep{li-etal-2022-neural}.

These works are valuable because better understanding how LLMs handle constructions could help us better understand what LLMs learn about linguistic structure \citep{baroni2021proper,linzen2021syntactic} and could also inform us as to what can be learned about language from primary data \citep{warstadt2022artificial,WarstadtB20}.
In our case, for an LLM to get the AANN construction right, a number of statistical regularities must be eschewed: ``a'' cannot be treated as a singular marker since the noun is plural, the normal ordering of the number and adjective must be reversed, and normal verb number agreement rules must in some cases be suspended.
Understanding whether these heuristics (which work well for the vast majority of text) can be overcome can guide us towards future work understanding \textit{how} they are overcome.

Here, I ask what GPT-3 \texttt{text-davinci-002} (now often classed as an instance of GPT-3.5) learns about the AANN construction by testing its sensitivity to several constraints proposed in the literature.
In doing so, I treat the LLM as a linguistic test subject \citep{linzen_assessing_2016,futrell-etal-2019-neural,wilcox-etal-2021-targeted,warstadt-etal-2019-neural,ettinger-2020-bert}.
I use a custom prompt to elicit quantitative grammaticality judgments \citep{t2016empirical,gibson_need_2013} from GPT-3,
and show that the prompt performs well on CoLA, a data set of binary acceptability judgments on a carefully constructed 10,657 English sentences \citep{warstadt-etal-2019-neural}.
I then unleash it on the AANN construction and compare GPT-3 to human ratings.

\begin{figure}[t]
    \centering
    \includegraphics[width=.8\columnwidth]{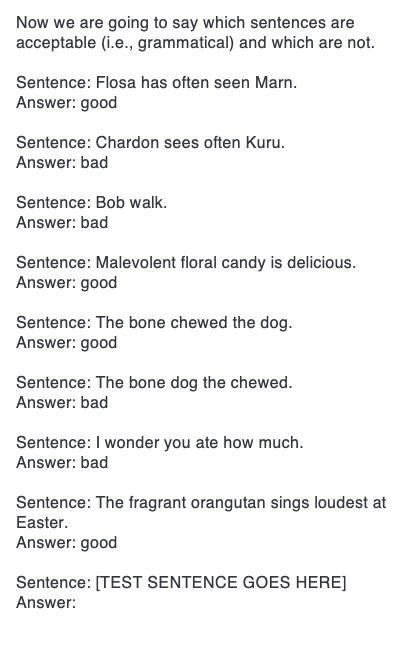}
    \caption{The prompt used for attaining grammaticality judgments. The target sentence is inserted, and then GPT-3 is asked to generate one token (overwhelmingly likely to be ``good'' or ``bad''). That token's probability is taken as a rating.}
    \label{fig:prompt}
\end{figure}
\section{Methods}

Attaining acceptability judgments from language models is not straightforward. 
Simply comparing sentence probabilities is difficult because they are dependent on the individual lexical items, as well as sentence length \citep{warstadt-etal-2020-blimp}.  
I rely on the prompting paradigm to elicit acceptability judgments---validating the measure first using a known data set of judgments from CoLA.
The prompt was created by drawing on a combination of CoLA training sentences and handcrafted sentences and iteratively experimenting. 
Ultimately, a diverse 8 sentences were chosen, some of which intentionally have low lexical probability to ensure that the model does not call all low-probability strings ungrammatical.
Each prompt example sentence appears along with a binary judgment: ``good'' or ``bad''.
Then, GPT-3 is passed the prompt, along with the critical test sentence, and asked to generate one more token (always either ``good'' or ``bad''). 
The probability of the generated word (whether ``good'' or ``bad'') is our numerical rating.

To validate the measure, I tested the final prompt on the CoLA dev set.
It attained accuracy of 84\%, with a Matthew's correlation coefficient of 0.63.
This is worse than, but in the ballpark of, human inter-annotator agreement on CoLA which is 86\% and .697, respectively \citep{warstadt-etal-2019-neural}.
It is also comparable to top performers on the GLUE leaderboard for the CoLA sub-task.

\begin{table}[t]
\scriptsize
\begin{tabular}{p{.15\linewidth}|p{.75\linewidth}}
\hline \bottomrule
\textbf{template} & \textbf{temporal}: The family spent X in London; The diplomat worked X in Nairobi; The tourist stayed X in Papua New Guinea; \textbf{objects}: She bought X; They discovered X; Someone saw X; \textbf{human}: We served dinner to X;  X greeted us at the door; We congratulated X; \textbf{art}: The newspaper reviewed X; I experienced X; Please enjoy X; \textbf{distance}: He drove X; Someone walked X; Someone traveled X; \textbf{unitlike}: Luis took in X; They consumed X; It lasted X \\ \hline
\textbf{template for agreement task} & \textbf{temporal}: X is/are just what you need; X is/are ideal \textbf{objects}: X is/are available; X make(s) a lovely gift; \textbf{human}: X regularly show(s) up at the door; X is/are here \textbf{art}: X was/were reviewed in the newspaper; X was/were enjoyed \textbf{distance}: X is/are a long way; X is/are not far; \textbf{unitlike} X was/were uncovered; X was/were make(s) an impression \\ \hline
\textbf{adj} & \textbf{ambig}: astonishing; incredible; impressive; disappointing; surprising; devastating; pathetic; remarkable; mediocre; unsatisfying; \textbf{qualitative}: lovely; beautiful; enchanting; soothing; charming; disgusting; uninviting; haunting; hideous; ugly; \textbf{quant}: mere; staggering; whopping; hefty; paltry; meager; extra; measly; substantial; record-setting; \textbf{stubborn}: large; big; small; round; tall; \textbf{color}: blue; green; red; yellow; orange; \textbf{human}: lucky; talented; graceful; fancy; friendly; collegial; hopeful; shy; bold; grinning \\  \hline
\textbf{noun} & \textbf{human}: soldiers; students; athletes; pianists; teammates; lawyers; doctors; actors; Americans; bankers; \textbf{objects}: desks; marbles; pencils; belts; forks; chairs; cans; bananas; apples; trays; \textbf{art}: movies; paintings; books; shows; operas; \textbf{temporal}: days; weeks; months; years; hours; \textbf{distance}: meters; feet; yards; blocks; steps; \textbf{unit_like}: pages; acts; paragraphs; awards; meals\\ \hline
\textbf{num.} & three; five; six; eight; ten; twenty; fifty; 500; 1000; 10,000; 21; 51; 512; 1,429; 21,234\\
\hline
\bottomrule
\end{tabular}
\caption{A superset of the items used, which were combined in various ways across experiments. In the templates, X is replaced by the AANN construction.}\label{tab:items}
\end{table}

I used this technique to test the AANN construction by templatically constructing sentences in which I parametrically vary the main sentence template, adjective, numeral, and nominal, from the superset shown in Table~\ref{tab:items}.
Templates were designed to work with the key manipulations. Certain nouns work with some templates and not others, to ensure template/~noun pairs are always plausible. 

Adjectives in the AANN construction behave differently depending on whether they are quantitative (i.e. modify the numeral  as in ``a mere 5 days''), qualitative (e.g., modify the noun as in ``a beautiful five days), or are ambiguous between the two (e.g., ``an astonishing five days'' which leaves it unclear whether the number of days is astonishing, or the days themselves). 
It has been claimed  \citep[e.g.,][]{dalrymple_amazing_2019,solt_two_2007,keenan_pleasant_2013} that quantitative and ambiguous adjectives are typically more acceptable than qualitative ones in AANN---although there are specific instances where qualitative adjectives are acceptable.
I also consider ``stubbornly distributive'' \citep{schwarzschild_stubborn_2011} adjectives (e.g., ``large'' or ``blue''), which ``stubbornly'' refer to individuals even when applied to a group. For instance, ``The chairs are large.'' can refer only to the individual chairs being large, not to the collective group of chairs being large.
It's claimed \citep{bylinina_zero_2018,ionin_cardinals_2018,keenan_pleasant_2013} that this same property makes a phrase like ``a large five trees'' less acceptable, compared to something like ``a beautiful five trees'' (in which it is possible for ``beautiful'' to refer to not just the individual trees but to the collection of trees).

\citet{solt_two_2007} and others observe measure nouns work best in the AANN construction, but that other nouns can be okay as long as they can be treated as a single unit. I sampled nouns from 6 categories, as shown in Table~\ref{tab:items}. Also as shown in that table, I sampled numerals of various kinds but focused on ``three'' and ``five'' for the human experiments and most analyses. See  Appendix~\ref{app:numerals} for a discussion of sensitivity to the numeral.

From these templates and candidate words, I generate semantically plausible sentences (meaning that,  throughout the experiments, I only use human-appropriate sentence templates with human nouns and object-appropriate ones with object nouns).
Depending on the specific question in each experiment, I run controlled subsets of these sentences (or their degenerate variants as in Experiment 1) through GPT-3 to obtain acceptability judgments, getting the probability of ``good'' or ``bad'' as the next word in the continuation.

I also use Amazon's Mechanical Turk to obtain human judgments on a subset of the test sentences. I asked raters to rate sentences on a rating bar scale from 1-10. For Experiments 1 and 3, each rater rated 3 critical sentences. They rated 18 critical sentences in Experiment 2 since there were many more conditions to test in that experiment.

To guard against raters becoming inured to the construction, half or more items were fillers not involving AANN.
To maintain similar calibration between humans and GPT-3, these fillers always included all example judgments used in the GPT-3 prompt.
I excluded participants who did the survey more than once, who did not rate the good filler items at least 1 point higher than the bad, or who did not have a US IP address.
I obtained annotations for only a sample of sentences rated by GPT-3. When applicable, analyses focus on the union of sentences rated by both humans and GPT-3.

\section{Exp. 1: AANN Fundamentals}

First, I tested the basics of the AANN construction as laid out in the literature by having GPT-3 and human raters rate the AANN construction (e.g., ``a beautiful five days'') vs. the default (``five beautiful days'') vs. a select four degenerate conditions: one with the order of the numeral and adjective switched, one with no modifier, one with a singular noun, and one with no article ``a''.
To generate examples in each condition, I crossed 3 temporal nouns (days, weeks, months), with a low numeral (three), one of 14 appropriate adjectives, and one of 3 templates.
For the resulting sentences, I attained human ratings from MTurk (after exclusions, 126 raters rating 3 sentences each, for 378 total ratings) and GPT-3 and focus on that subset for analysis.

Figure~\ref{fig:fig1} shows results for this experiment. Although GPT-3 uses a wider range of the scale, both rate the AANN construction as just as good as the default and give lower ratings to the 4 degenerate versions.
Humans rate the 4 degenerate constructions as about equally bad (all between .46 and .53), whereas GPT-3 rates the versions with swapped adjective/numeral order and a missing article as significantly better than the versions with a missing plural and a missing modifier.
Running a mixed effect regression \citep{bates_fitting_2015,barr_random_2013} comparing the AANN construction to the ``default'' construction and the degenerate alternatives (treating the default construction as the baseline, with random intercepts for adjective class, adjective, and template; and, for humans, for rater), both GPT-3 and humans show no significant difference in rating between AANN and default (both $p > .05$), but do show a significant difference between AANN and all 4 degenerate conditions (all $ps < .0001$). See Appendix~\ref{app:default} for regression details.
Overall, I conclude that humans and GPT-3 ``get'' the AANN construction, even though they differ in the relative ratings of the bad variants.

\section{Exp. 2: Adjectives and nouns} 
In this experiment, I focus on only the AANN construction and parametrically vary the kinds of adjectives (quantitative, ambiguously quantitative/qualitative, qualitative, human-referring, color adjective, stubbornly distributive; see Table 1 for examples) and kinds of nouns in the sentences (art, distance, human nouns, object nouns, temporal nouns, unit-like nouns; again see Table 1).
This process produced a carefully controlled 12,960 unique sentences, from which we sampled a random subset for human ratings. 
After exclusions, there were 190 raters left who each rated 18 sentences, giving us ratings for 3,420 sentences.

\begin{figure}[t]
     \centering
    \includegraphics[width=1\columnwidth]{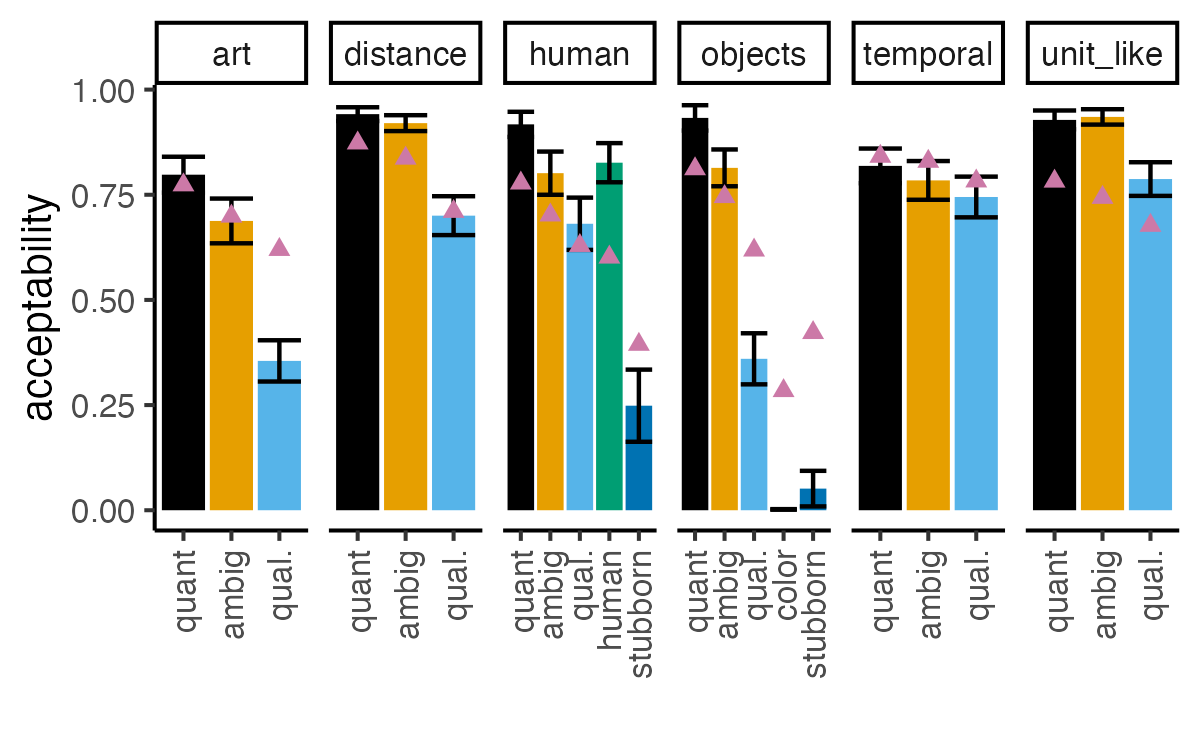}
    \vspace{-.3in}
\caption{GPT-3 acceptability scores broken down by adjective type (x-axis), and noun type (on facets). Human ratings are pink triangles.
}\label{fig:adjs}
     \end{figure}

I test whether GPT-3 and human raters  agree with the attested claims that (a) more measure-like nouns (e.g., temporal nouns, distance nouns, and unit-like nouns) are more acceptable in AANN, (b) qualitative adjectives are acceptable in only some AANN cases, and (c) stubbornly distributive adjectives (including color words) are not acceptable in the AANN construction.
To assess significance, I predicted acceptability separately for humans and GPT-3 in a mixed effect regression, using adjective class, noun class, and their interaction as predictors and with random effects for adjective, noun, numeral, and template.
I treated qualitative adjectives with temporal nouns as the baseline.

For both humans and GPT-3, there are significant differences in how nouns interact with adjectives 
(Figure~\ref{fig:adjs}).
For temporal, distance, and unit-like nouns, all adjective types show high acceptability (although the qualitative adjectives are rated lowest). 
For art and object nouns, qualitative nouns score significantly worse ($p<.01$ for both humans and GPT-3) than ambiguous or quantitative nouns (an observation consistent with the literature).
As predicted, colors and other stubbornly distributive adjectives (which can be tested only for humans and objects) show the lowest acceptability (significantly lower, $p<.01$, compared to qualitative adjectives, for both human and GPT-3 ratings).
See Appendix~\ref{app:adj} for regression details.

\section{Exp. 3: Adjective Order}

It is claimed that, in AANN, qualitative adjectives must appear before quantitative ones in order to be acceptable \citep{solt_two_2007}: ``The family spent a beautiful mere five days in London.''  is preferred over ``The family spent a mere beautiful five days in London.''.
To compare whether there is an effect of adjective ordering (putting the qualitative adjective before the quantitative one or vice versa), I ran an experiment crossing 3 templates; 5 adjectives (astonishing, impressive, beautiful, hideous, ugly); the noun ``days''; and the numeral ``three'' or ``five''.
I ran each sentence in two conditions (quantitative adjective first or qualitative adjective first).
For instance, I compared: ``The family spent a beautiful mere five days in London.'' to ``The family spent a mere beautiful five days in London.''
This left 60 sentences total, rated by 99 raters (each sentence rated between 18 and 36 times; 1,782 ratings total).

GPT-3 significantly prefers the order \textit{dispreferred} in the literature (quantitative first, as in ``a mere beautiful five days''; $\beta = .04, p < .01$). 
Humans (n=99) showed no clear preference according to the model, under the same analysis (but with a random intercept for rater): $\beta = -.513, p > .05$.
Thus, the attested claim does not replicate.
But the ratings for these sentences are relatively low overall, so we should remain open to the possibility that there are better examples of the double-adjective constructions than the ones tested. See Appendix~\ref{app:order} for regression details.

\begin{figure}[t]
         \centering
         \includegraphics[width=\columnwidth]{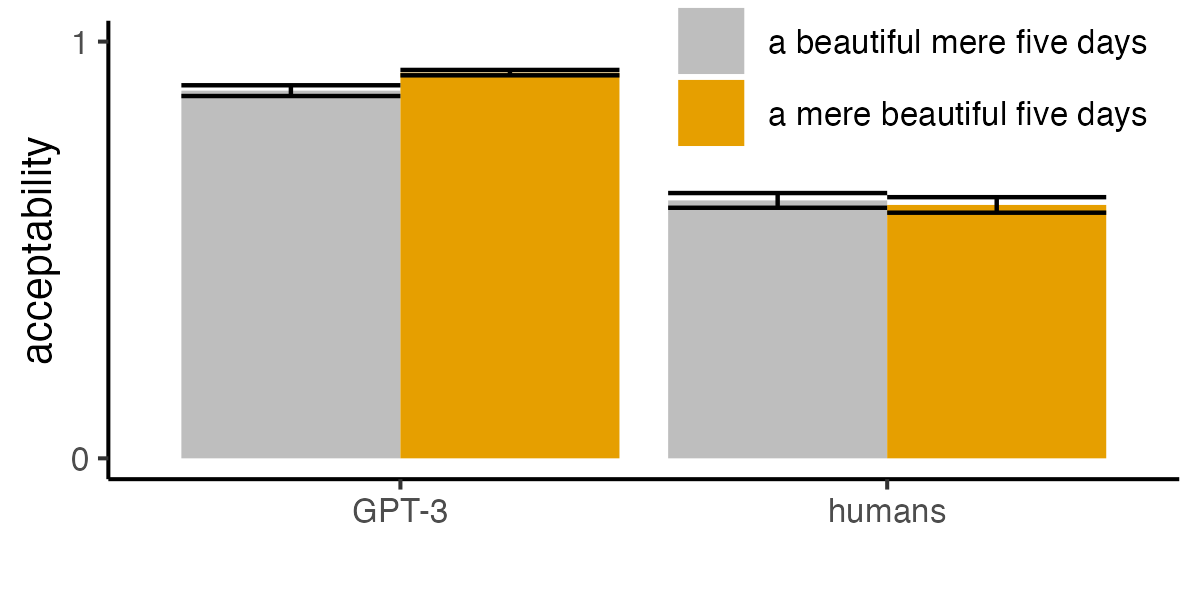}
         \caption{GPT-3 and human preference for adjective order (quantitative before qualitative; qualitative before quantitative).}
         \label{fig:twoadjs}
\end{figure}

\section{Exp. 4: Verb Agreement}

The AANN construction also challenges number agreement.
AANN subjects sometimes take singular verbs (when the noun phrase would be singular anyway, as in ``A mere fifty cents for a cup of coffee sounds/*sound reasonable to me!''); sometimes plural (``A delicious four courses *was/were served in the main dining room.''), and sometimes either (``A healthy two runs weekly was/were prescribed by the doctor.'' See \citet{keenan_pleasant_2013}.

\begin{figure}[t]     
         \centering
    \includegraphics[width=.9\columnwidth]{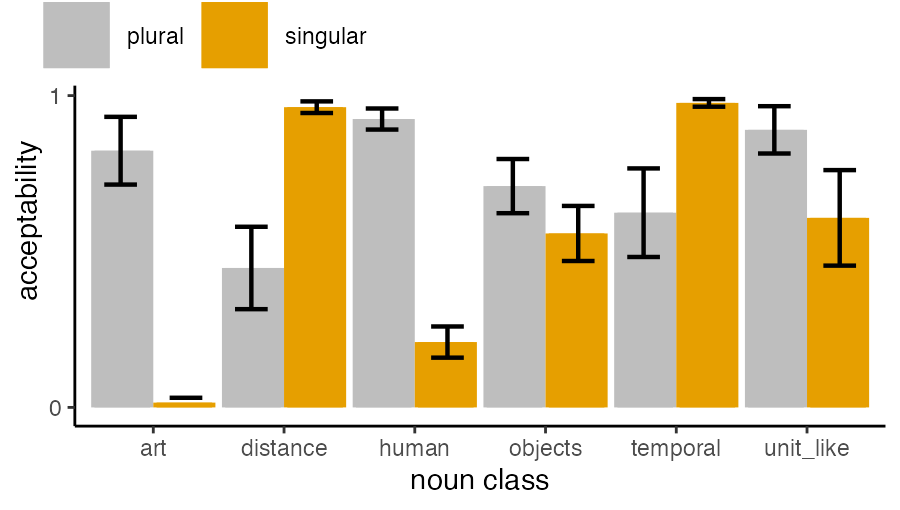}
         \caption{Mean GPT-3 acceptability ratings in the AANN construction for plural and singular verb agreement, as a function of noun class.
         }\label{fig:agree}
\end{figure}

I tested agreement by comparing phrases which differed only in the verb number (e.g., comparing ``A beautiful five days is...'' vs. ``A beautiful five days are....''). 
Sampling a subset of noun classes (art, distance, human nouns, objects, unit-like nouns, and temporal nouns), I generated a total of 280 sentences (each appearing with a singular or plural verb for a total of 560 sentences) and attained judgments from GPT-3.
As shown in Figure~\ref{fig:agree}, these results replicated in detail several attested judgments in the literature: art nouns, human nouns, unit-like nouns, and object nouns all prefer the plural for the AANN construction (art and humans almost categorically so). 
Distance nouns and temporal nouns prefer the singular (although temporal nouns can also take the plural).
See Appendix~\ref{app:agree} for details.

\section{Conclusion}

This work shows that GPT-3 can recognize and use the \textit{form} of the AANN construction in a relatively (but not perfectly) human-like way, matching judgments across a variety of conditions, which is not the same thing as showing that it understands the meaning or \textit{function} of the construction \citep{mahowald2023dissociating}.
In future work, we should study not just the form of the construction but its \textit{construal} \citep{trott-etal-2020-construing}.
This may be particularly relevant since \citet{weissweiler2022better}, which studies the ``Xer the Yer'' construction (e.g., ``the better the criticism, the better the science'' [COCA]), show that LLMs can recognize the construction but fail at tests of understanding its meaning.

That said, GPT-3's performance on the AANN construction demands a significant amount of constructional knowledge and involves overriding major widespread ``rules'' of grammar (e.g., that the article \textit{a} signals a singular noun). 
Future work exploring how LLMs override those heuristics, perhaps using causal intervention techniques \citep[][]{ravfogel-etal-2021-counterfactual,geiger2021causal} could illuminate their syntactic processing mechanisms.

\section{Limitations}

Many researchers have pointed out that the AANN construction is sensitive to context. For instance, \citet{solt_two_2007} points out that ``a hungry thirty hikers'' may be acceptable in some sentences (namely ones where ``hikers'' is more easily construed as a single unit) than others. 
Because of the cost of running sentences through GPT-3 and on MTurk and the combinatoric nature of the construction, I could only run a constrained number of templates and did not consider larger context. 
Broader context may matter and so these results should be taken to apply to the particular contexts shown, which is why I show the exact templates used in the main text.

Another limitation is that the task of how to prompt GPT-3 for grammaticality judgments is not a settled question.
Our focus in this paper was not on settling it, so I used one particular method for prompting GPT-3.
While I explored the prompt space some, it is likely there are better prompts out there that would make GPT-3's performance on the task better.
It's also possible that prompting GPT-3 for grammaticality judgments is not the best way to ascertain its knowledge of language and that a more naturalistic task would produce different results.

As has often been pointed out in the linguistics literature, naive human judges of out-of-context sentences may sometimes tap into different processes than they would when encountering language in the wild. 
Moreover, English is not a monolith and this construction's acceptability may vary across dialects of English.
In a more detailed human study, it would be possible to tease apart effects of different dialects on ratings.

GPT-3 \texttt{text-davinci-002} is often categorized as GPT-3.5 because it is trained on more than just a word prediction task, and so we should not interpret its output as being purely reflective of what is learned by word prediction alone.

Finally, I note that I use templatically constructed sentences, which differ in important ways from naturalistic ones.

\section{Acknowledgments}

I acknowledge funding from NSF Grant 2104995.
For helpful comments, I thank Robbie Kubala, Adele Goldberg, Gabriella Chronis, Katrin Erk, Alex Warstadt, Steve Wechsler, Liam Blything, attendees of the Princeton language group meeting, and an anonymous three reviewers.

\bibliography{anthology,custom}
\bibliographystyle{acl_natbib}

\appendix

\section{Frequency of AANN}
\label{app:freq}

I sampled the AANN construction on SketchEngine for the English Web 2020 corpus with the following prompts:
\texttt{[lemma=''a[n]*''] [tag="JJ.*"] [tag="CD.*"] [tag="NNS"]} for AANN and \texttt{[tag="CD.*"] [tag="JJ.*"] [tag="NNS"]} for the vanilla construction. 
The AANN showed up 23.62 per million tokens, compared to 457.22 for the vanilla construction.
Of the AANN construction examples, the vast majority contain quantitative adjectives (e.g., \textit{mere}, \textit{staggering}, etc.) and measurement nouns.
Of a sample of 200 AANN constructions that I manually inspected on SketchEngine, none contained a qualitative adjective.

\section{Numerals}
\label{app:numerals}

I sampled numerals from round low numbers (``three'', ``five'', ``six'', ``eight'', ``ten''), medium numbers (``twenty'', ``fifty''), high numbers (500, 1000, 10,000), non-rounded medium numbers (21 and 51), and non-rounded high numbers (1,429 and 21,234). I focused on round low numbers (``three'' and ``five'') for most analyses. 

For the overall analysis (across all adjectives and nouns in the main experiment), I get the below AANN ratings from GPT-3.
\begin{table}[ht]
\centering
\small
\begin{tabular}{llr}
  \hline
\textbf{numclass} & \textbf{example} & \textbf{avg. AANN score} \\ 
  \hline
num-high & 500 & 0.68 \\ 
  num-high\_odd & 1,429 & 0.93 \\ 
  num-low & three & 0.80 \\ 
  num-med & twenty & 0.70 \\ 
  num-med\_odd & 21 & 0.73 \\ 
   \hline
\end{tabular}
\end{table}

Low numbers like ``three'' and ``five'' are rated higher somewhat than other numbers, with the exception of high, non-round numbers (e.g., 1,429). 
These are rated unusually highly.
Because of this anomalous behavior, I focus mostly on the low numerals for the analysis and did not attain human ratings for these other numerals.
It remains an open question how humans would rate a sentence like ``We spent a beautiful 1,429 days in London.''

\section{AANN vs. default vs. degenerate regression}\label{app:default}

I run the following regression using the R \texttt{lme4} \citep{bates_fitting_2015} package.

\begin{verbatim}
    lmer(value ~ construction +
               (1|adjclass) +
               (1|adj) +
               (1 | temp))
\end{verbatim}

\noindent where I treat the construction as a predictor (with the AANN constructor as a default) and adjclass, adjective, template as random effects. (Other random effects were removed to help the model converge, by iteratively removing ones with the smallest variance.)

For humans, there is also a random intercept for worker.

\begin{table}[h]
\centering
\small
\begin{tabular}{rrrrr}
  \hline
 coef. & $\beta_{gpt3}$  & Sig & $\beta_{human}$  & Sig  \\ 
  \hline
(Intercept) & 0.99 & * & .81 & * \\ 
  five ADJ days & 0.00 & & .06 &  \\ 
  a five ADJ days & -0.22 & * & -.39 & *\\ 
  a five days & -0.87  & * & -.35 & *\\ 
  a ADJ five day & -0.72 & * & -.35 & *\\ 
  ADJ five days & -0.20 & * & -.38  & *\\ 
   \hline
\end{tabular}
\caption{Fixed effect coefficients for GPT-3 comparing across constructions on the subset of sentences also run with human annotators}
\end{table}

\section{Regression for adjective x noun sub-experiment with GPT-3 and humans}\label{app:adj}

To assess significance, I predicted acceptability separately for humans and GPT-3 in a mixed effect regression, using adjective class, noun class, and their interaction as predictors and with random effects for adjective, noun, numeral, and template.
I treated qualitative adjectives with temporal nouns as the baseline.
For the human regression, it was identical except I included a random intercept for each worker, with a random slope for adjective class and noun class (but not their interaction, due to convergence issues).

\begin{verbatim}
    lmer(rating ~ adjclass * 
           nounclass + 
           (1|adj) + 
           (1|noun) + 
           (1|num) + 
           (1 |template))
\end{verbatim}

\begin{table}[ht]
\small
\centering
\begin{tabular}{rrrl}
  \hline
 & beta & t-value & p$<$.05 \\ 
  \hline
(Intercept) & 0.80 & 12.09 & * \\ 
  adj-quant & 0.02 & 0.27 &  \\ 
  adj-stubborn & -0.62 & -10.27 & * \\ 
  adj-ambig & -0.02 & -0.32 &  \\ 
  noun-unit\_like & 0.14 & 3.49 & * \\ 
  noun-objects & 0.02 & 0.58 &  \\ 
  noun-human & 0.03 & 0.76 &  \\ 
  noun-distance & 0.14 & 3.40 & * \\ 
  noun-art & -0.10 & -2.33 & * \\ 
  adj-quant:noun-unit\_like & -0.02 & -0.80 &  \\ 
  adj-quant:noun-objects & 0.09 & 2.63 & * \\ 
  adj-stubborn:noun-objects & -0.17 & -3.64 & * \\ 
  adj-quant:noun-human & 0.08 & 2.24 & * \\ 
  adj-quant:noun-distance & -0.01 & -0.28 &  \\ 
  adj-quant:noun-art & 0.07 & 2.35 & * \\ 
   \hline
\end{tabular}
\caption{Fixed effect coefficients for GPT-3 comparing the adjective class x noun class manipulation.}
\end{table}

\begin{table}[ht]
\small
\centering
\begin{tabular}{rrrl}
  \hline
 & beta & t-value & p$<$.05 \\ 
  \hline
(Intercept) & 0.73 & 16.84 & * \\ 
  adj-quant & 0.10 & 2.63 & * \\ 
  adj-stubborn & -0.25 & -5.58 & * \\ 
  adj-ambig & 0.09 & 2.65 & * \\ 
  noun-unit\_like & -0.08 & -3.11 & * \\ 
  noun-objects & -0.09 & -3.38 & * \\ 
  noun-human & -0.12 & -4.25 & * \\ 
  noun-distance & 0.00 & 0.17 &  \\ 
  noun-art & -0.12 & -4.28 & * \\ 
  adj-quant:noun-unit\_like & 0.02 & 0.72 &  \\ 
  adj-quant:noun-objects & 0.06 & 2.04 & * \\ 
  adj-stubborn:noun-objects & -0.04 & -0.91 &  \\ 
  adj-quant:noun-human & 0.05 & 1.62 &  \\ 
  adj-quant:noun-distance & 0.03 & 1.27 &  \\ 
  adj-quant:noun-art & 0.04 & 1.51 &  \\ 
   \hline
\end{tabular}
\caption{Fixed effect coefficients for human annotators comparing the adjective class x noun class manipulation.}
\end{table}

\section{Regression for adjective ordering}\label{app:order}

I ran a mixed effect linear regression predicting the GPT-3 score from the condition (quantitative-first vs. qualitative-first), with random intercepts for adjective, numeral, and template.

\begin{verbatim}
    l = lmer(value ~ cond + 
            (1|adj)  +
           (1|num) + 
           (1 |template)
\end{verbatim}

\section{Agreement}\label{app:agree}

I ran a regression predicting the rating based on the nounclass, and its interaction with whether there was singular plural. I included random intercepts for noun, adjective, and template, with a random slope for whether the verb was singular or plural on the noun factor.

\begin{verbatim}
     lmer(rating ~ singplur * 
           nounclass +
           (1 + singplur|noun) + 
           (1|adj) + 
           (1|template))
\end{verbatim}

Results appear in Table~\ref{tab:agree}, where the baseline values are temporal nouns in the plural (e.g., ``days''). Singular verbs are preferred overall, relative to plurals for the temproal nouns (main effect of ``singular''). There are various main effects of nouns, but the critical effects here are interactions. 
There are significant effects such that, \textit{relative to temporal nouns} (e.g., ``days''), unit-like nouns are less likely to prefer singular agreement, object nouns are less likely to prefer singular agreement, human nouns are less likely to prefer singular agreement, and art nouns are less likely to prefer singular agreement. Distance nouns are more likely than temporal nouns to prefer singular agreement (but not significantly so). 

\begin{table}[t]
\centering
\small
\begin{tabular}{rrrl}
  \hline
 & beta & t-value & p$<$.05 \\ 
  \hline
(Intercept) & 0.57 & 4.88 & * \\ 
  singular & 0.35 & 3.14 & * \\ 
  noun-unit\_like & 0.27 & 3.37 & * \\ 
  noun-objects & 0.17 & 2.50 & * \\ 
  noun-human & 0.29 & 4.27 & * \\ 
  noun-distance & -0.18 & -2.25 & * \\ 
  noun-art & 0.20 & 2.52 & * \\ 
  singular:noun-unit\_like & -0.63 & -6.40 & * \\ 
  singular:noun-objects & -0.50 & -6.12 & * \\ 
  singular:noun-human & -1.07 & -12.96 & * \\ 
  singular:noun-distance & 0.16 & 1.65 &  \\ 
  singular:noun-art & -1.16 & -11.71 & * \\ 
   \hline
\end{tabular}
\caption{Fixed effect coefficients and significance values for an experiment comparing whether nouns in the AANN construction prefer singular or plural verbs.}
\label{tab:agree}
\end{table}

\end{document}